\documentclass[10pt,twocolumn,letterpaper]{article}

\usepackage[pagenumbers]{cvpr} 

\usepackage{graphicx}
\usepackage{amsmath}
\usepackage{amssymb}
\usepackage{booktabs}
\usepackage{mathtools}
\usepackage[numbers,sort,compress]{natbib}

\usepackage[symbol*]{footmisc}

\usepackage[export]{adjustbox}
\usepackage{multirow}
\usepackage[ruled,vlined]{algorithm2e}
\usepackage[inline]{enumitem}
\usepackage{nicefrac}

\usepackage{ifthen}
\usepackage[normalem]{ulem}
\usepackage{xcolor}

\usepackage{xspace}
\newcommand{\hpvaegan}{HP-VAE-GAN\xspace}

\usepackage[pagebackref,breaklinks,colorlinks]{hyperref}

\usepackage[capitalize]{cleveref}
\crefname{section}{Sec.}{Secs.}
\Crefname{section}{Section}{Sections}
\Crefname{table}{Table}{Tables}
\crefname{table}{Tab.}{Tabs.}

\def\cvprPaperID{14} 
\def\confName{CVFAD}
\def\confYear{2022}

\begin{document}

\title{\vspace{-1cm}Diverse Video Generation from a Single Video\vspace{-.2cm}}

\author{Niv Haim\thanks{Equal contribution}, Ben Feinstein\footnotemark[1], Niv Granot, Assaf Shocher, Shai Bagon, Tali Dekel and Michal Irani \\
Weizmann Institute of Science, Rehovot, Israel}

\maketitle

\begin{figure}
    \centering
    \includegraphics[width=\columnwidth]{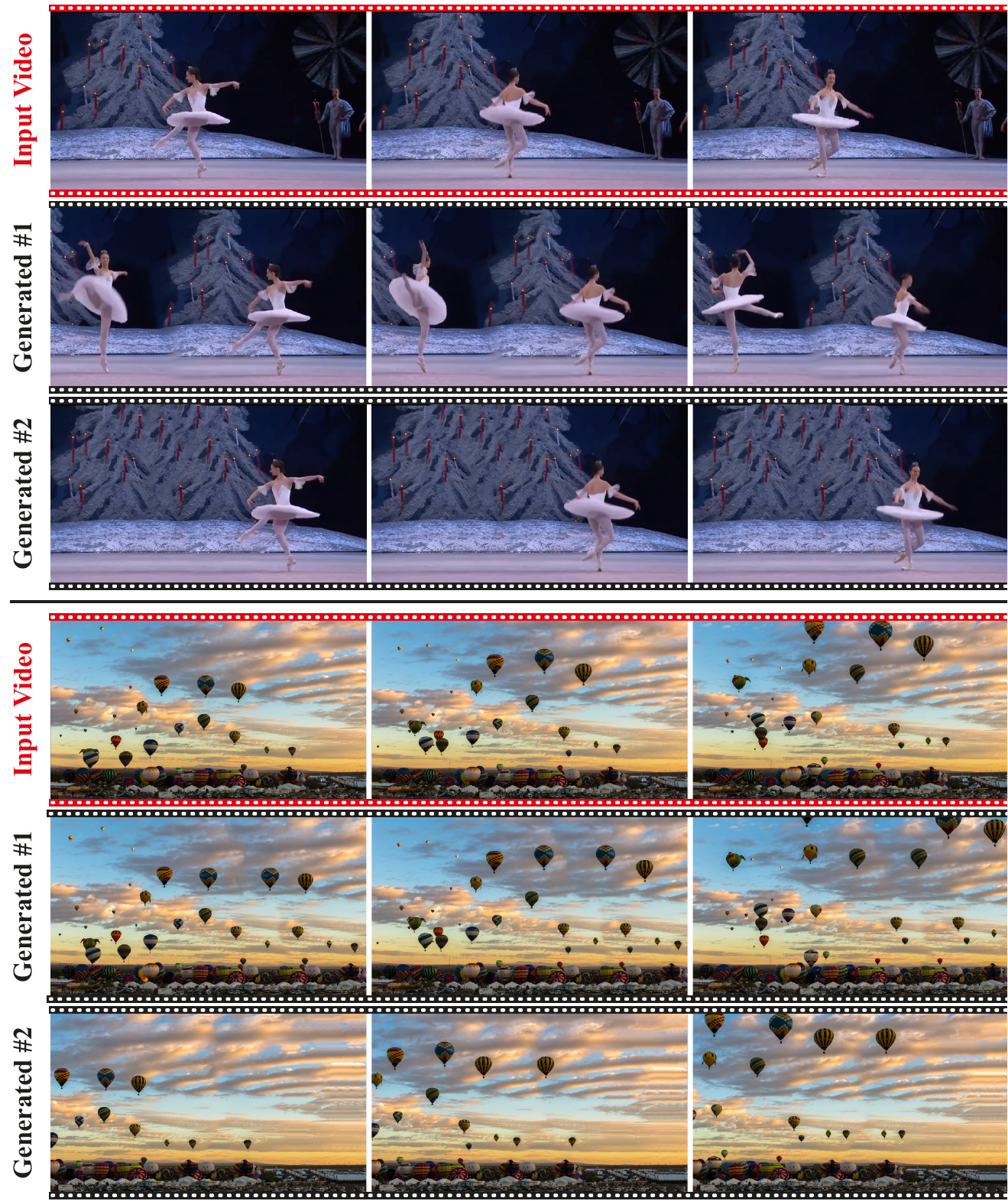}
    \caption{
    Given an input video (red), we generate similarly looking videos (black) capturing both appearance of objects as well as their dynamics. The diversity of the outputs is both spatially (e.g., number of dancers and their positions are different from the input video) and temporally (generated dancers are not synced). \newline As we present video results, the reader is encouraged to \textbf{check our supplementary material}
    \vspace{-.5cm}
    }
    \label{fig:generation_examples}
\end{figure}

GANs are able to perform generation and manipulation tasks, trained on a single video. However, these single video GANs require unreasonable amount of time to train on a single video, rendering them almost impractical. 
In this paper we question the necessity of a GAN for generation from a single video, and introduce a non-parametric baseline for a variety of generation and manipulation tasks.
We revive classical space-time patches-nearest-neighbors approaches and adapt them to a scalable unconditional generative model, without any learning.
This simple baseline surprisingly outperforms single-video GANs in visual quality and realism (confirmed by quantitative and qualitative evaluations), and is disproportionately faster (runtime reduced from several days to seconds). Our approach is easily scaled to Full-HD videos.
We also use the same framework to demonstrate video analogies and spatio-temporal retargeting. 
These observations show that classical approaches significantly outperform heavy deep learning machinery for these tasks. This sets a new baseline for single-video generation and manipulation tasks, and no less important --  makes diverse generation from a single video practically possible for the first time.
%

\vspace{-4pt}
\section{Introduction}
\vspace{-4pt}
\label{sec:intro}

Generation and editing of natural videos remain challenging, mainly due to their large dimensionality and the enormous space of motion they span. 
Most modern frameworks train generative models on a large collection of videos, producing high quality results for only a limited class of videos. These include extensions of GANs~\cite{goodfellow2014generative} to video data~\cite{vondrick2016generating, tulyakov2018mocogan, aigner2018futuregan, wang2021inmodegan, saito2017temporal,lee2018stochastic} and video-to-video translation~\cite{wang2018video, wang2019few, mallya2020world, wang2020imaginator, chan2019everybody, yang2020transmomo,bansal2018recycle}, autoregressive sequence prediction~\cite{ballas2015delving,villegas2017decomposing, babaeizadeh2017stochastic, denton2018stochastic, villegas2018hierarchical, villegas2019high, aksan2019stcn, franceschi2020stochastic} and more. 
While externally-trained generative models produce impressive results, they are restricted to the types of video dynamics in their training set. 
On the other side of the spectrum are \emph{single-video GANs}.
These video generative models  train on a \emph{single} input video,
learn its distribution of space-time patches, and are then able to generate a diversity of new videos with the same patch distribution~\cite{gur2020hierarchical, arora2021singan}. 
However, these take very long time to train for each input video, making them applicable to only small spatial resolutions and to very short videos (typically, very few small frames). Furthermore, their output oftentimes shows poor visual quality and noticeable visual artifacts. These shortcomings render existing single-video GANs impractical and unscalable.

Video synthesis and manipulation of a single video sequence 
based on its distribution of space-time patches 
dates back to classical  pre-deep learning methods. 
These classical methods 
demonstrated impressive results for various applications, such as video retargeting~\cite{simakov2008summarizing,rubinstein2008improved,wolf2007videoretarget,krahenbuhl2009disney}, video completion~\cite{wexler2004space,liu2009video,huang2016temporally}, video texture synthesis~\cite{kwatra2003graphcut,kwatra2005texture,kwatra2007texturing,bhat2004flow,fivser2016stylit, jamrivska2015lazyfluids} and more.
With the rise of deep-learning, these methods gradually, perhaps unjustifiably, became less popular. 
Recently, \citet{granot2021drop} revived classical patch-based approaches for image synthesis, and was shown to significantly outperform \emph{single-image} GANs in both run-time and visual quality.

In light of the above-mentioned deficiencies of single-video GANs, and inspired by~\cite{granot2021drop}, 
we propose a fast and practical method for video generation from a single video that we term VGPNN (\emph{Video Generative Patch Nearest Neighbors}).
In order to handle the huge amounts of space-time patches in a single video sequence, we use the classical fast approximate nearest neighbor search method PatchMatch by ~\citet{barnes2009patchmatch}.
By adding stochastic noise to the process, our approach can generate a large diversity of random different video outputs from a single input video in an unconditional manner.

Like single-video GANs, our approach enables the diverse and random generation of videos. However, in contrast to existing single-video GANs, we can generate \emph{high resolution} videos, while reducing runtime by many orders of magnitude, thus making diverse unconditional video generation from a single video realistically possible for the first time. 

In addition to diverse generation from a single video, by employing robust optical-flow based descriptors we use our framework to transfer the dynamics and motions between two videos with different appearance (which we call ``video analogies''). We also show the applicability of our framework to spatio-temporal video retargeting and to conditional video inpainting.

\noindent To summarize, our contributions are as follows:
\begin{itemize} [topsep=0pt,itemsep=-1ex,partopsep=1ex,parsep=1ex,leftmargin=*]
\item We show that our space-time patch nearest-neighbors approach, despite its simplicity, outperforms single-video GANs by a large margin, both in runtime and in quality.
\item Our approach is the first to generate diverse high resolution videos (spatial or temporal) from a single video.
\item We demonstrate the applicability of our framework to other applications: video analogies, sketch-to-video, spatio-temporal video retargeting and conditional video inpainting.
\end{itemize}

Our code and data will be released.

\vspace{-8pt}
\section{Method} 
\vspace{-8pt}
\label{sec:method}

\begin{figure}
    \includegraphics[width=\columnwidth]{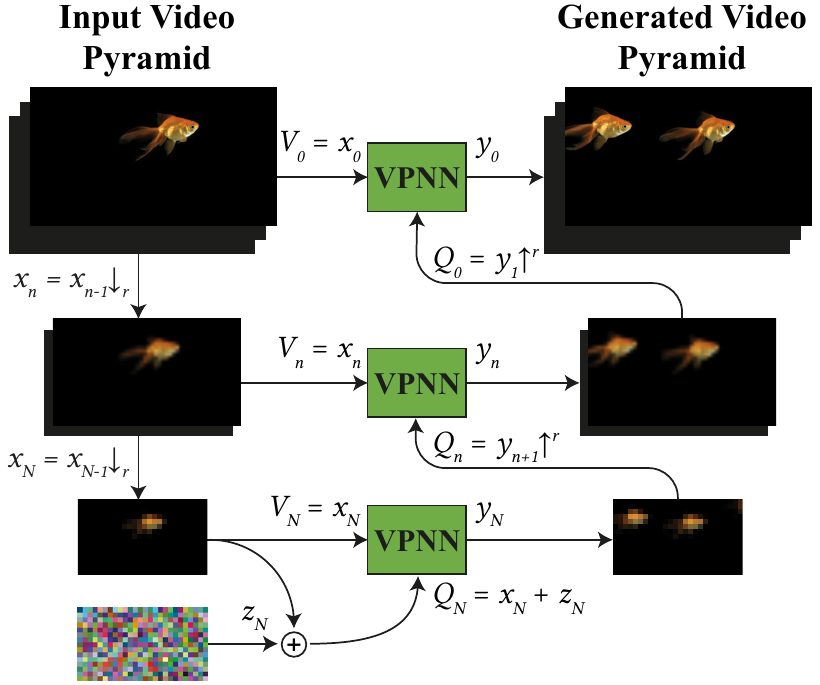}
    \caption{\textbf{VGPNN Architecture:} Given a single input video $x_0$, a spatio-temporal pyramid is constructed and an output video $y_0$ is generated coarse-to-fine. At each scale, VPNN module (Fig.~\ref{fig:method_vpnn}) is applied to transfer an initial guess $Q_n$ to the output $y_n$ which shares the same space-time patch distribution as the input $x_n$. At the coarsest scale, noise is injected to induce randomness.
    \vspace{-16pt}
    }
    \label{fig:method_pyramid}
\end{figure}

Our main task is to generate diverse video samples based on a single natural input video, such that the generated outputs have similar appearance and motions as the original input video, but are also visually different from one another.

In order to capture both spatial and temporal information of a single video, we start by building a spatio-temporal pyramid and operate coarse-to-fine to capture the internal statistics of the input video at multiple scales.
At each scale we employ a Video-Patch-Nearest-Neighbor module (\textit{VPNN}); VGPNN is in fact a sequence of VPNN layers. The inputs to each layer depend on the application, where we first focus on our main application of diverse video generation.

Given an input video $x$, we construct a spatio-temporal pyramid $\left\{x_0 \dots, x_N\right\}$, where $x_0=x$, and $x_n = x_{n-1} {\downarrow_{r}}$ is a bicubically downscaled version of $x_{n-1}$ by factor $r$ (${r=(r_H,r_W,r_T)}$, where $r_H=r_W$ are the spatial factors and $r_T$ is the temporal factor, which can be different). 

\paragraph{Multi-scale approach (Fig.~\ref{fig:method_pyramid}):}

At the coarsest level, the input to the first VPNN layer is an initial coarse guess of the output video. This is created by adding random Gaussian noise $z_N$ to $x_N$.
The noise $z_N$ promotes high diversity in the generated output samples from the single input. The global structure (e.g., a head is above the body) and global motion (e.g., humans walk forward), is prompted by $x_N$, where such structure and motion can be captured by \emph{small space-time} patches. 

Each space-time patch of the initial coarse guess ($x_N + z_N$) is then replaced with its nearest neighbor patch from the corresponding coarse input $x_N$. The coarsest-level output $y_N$ is generated by choosing at each space-time position the median of all suggestions from neighboring patches (known as ``voting'' or ``folding'').

At each subsequent level, the input to the VPNN layer is the 
upscaled output of the previous layer (${y_{n+1}\uparrow}^r$). 
Each space-time patch is replaced with its nearest neighbor patch from the corresponding input $x_n$ (using the same patch-size as before, now capturing finer details).
This way, the output $y_n$ in each level is similar in structure and in motion to the initial guess, but contains the same space-time patch statistics of the corresponding input $x_n$. The output $y_n$ is generated by median voting as described above.

To further improve the quality and sharpness of the generated output at each pyramid level ($y_n$), we iterate several times through the current level, each time using the current output $y_n$ as input to the current VPNN layer (similar to the EM-like approach employed in many patch-based works \citep[e.g.,][]{granot2021drop,simakov2008summarizing,barnes2009patchmatch,wexler2004space}).

\begin{figure}
    \includegraphics[width=\columnwidth]{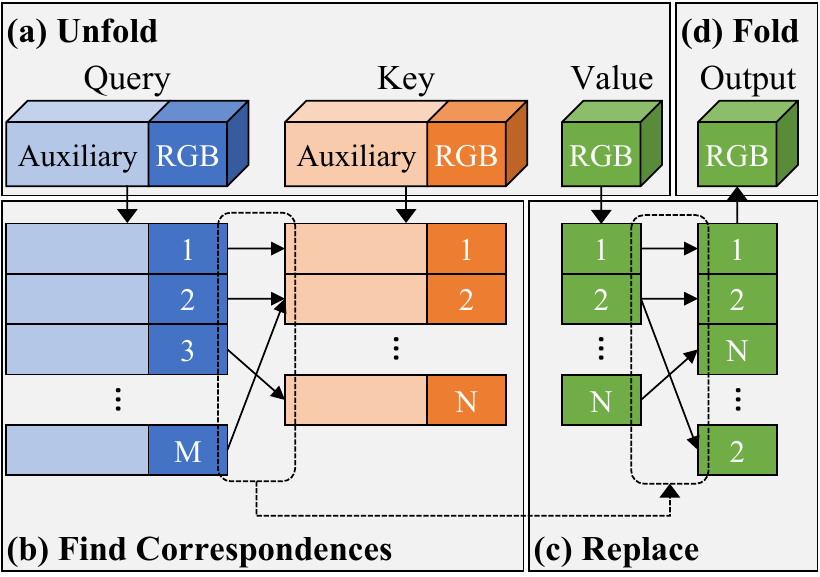}
  \caption{\textbf{VPNN module} 
  gets as input RGB videos query, key and value (QKV respectively). Q and K can be concatenated to additional auxiliary channels. It outputs an RGB video. (a) Inputs are unfolded to patches (each position now holds a concatenation of neighboring positions); (b) Each patch in Q finds its nearest neighbor patch in K. This is achieved by solving the NNF using PatchMatch \cite{barnes2009patchmatch}; (c) Each patch in Q is replaced with a patch from V, according to the correspondences found in stage (b); (d) Resulting patches are folded back to an RGB video output.
  \vspace{-.5cm}
  }
  \label{fig:method_vpnn}
\end{figure}

\paragraph{QKV scheme:} In several cases it is necessary to compare patches in another search space than the original RGB input space. To this end we adopt a QKV scheme (query, key and value, respectively) as used by \cite{granot2021drop}. For example, when comparing the upscaled output of previous layer to the corresponding level from the pyramid of the original video, the patches of the latter are sharper than the former. This is mitigated by setting $V=x_n$ and $K=x_{n+1}{\uparrow^r}$ which has a similar degree of blur as $Q={y_{n+1}\uparrow}$ where. Each patch $Q_i$ with nearest neighbour $K_j$ is replaced with $V_j$ ($i,j$ are spatio-temporal positions).
The QKV scheme is especially important in our video analogies application where it is used to include additional temporal information in the queries and the keys.

\vspace{-6pt}
\paragraph{Finding Correspondences:} 
\label{sec:patchmatch} 

\begin{figure}
    \centering
    \begin{tabular}{c}
        \includegraphics[width=\columnwidth]{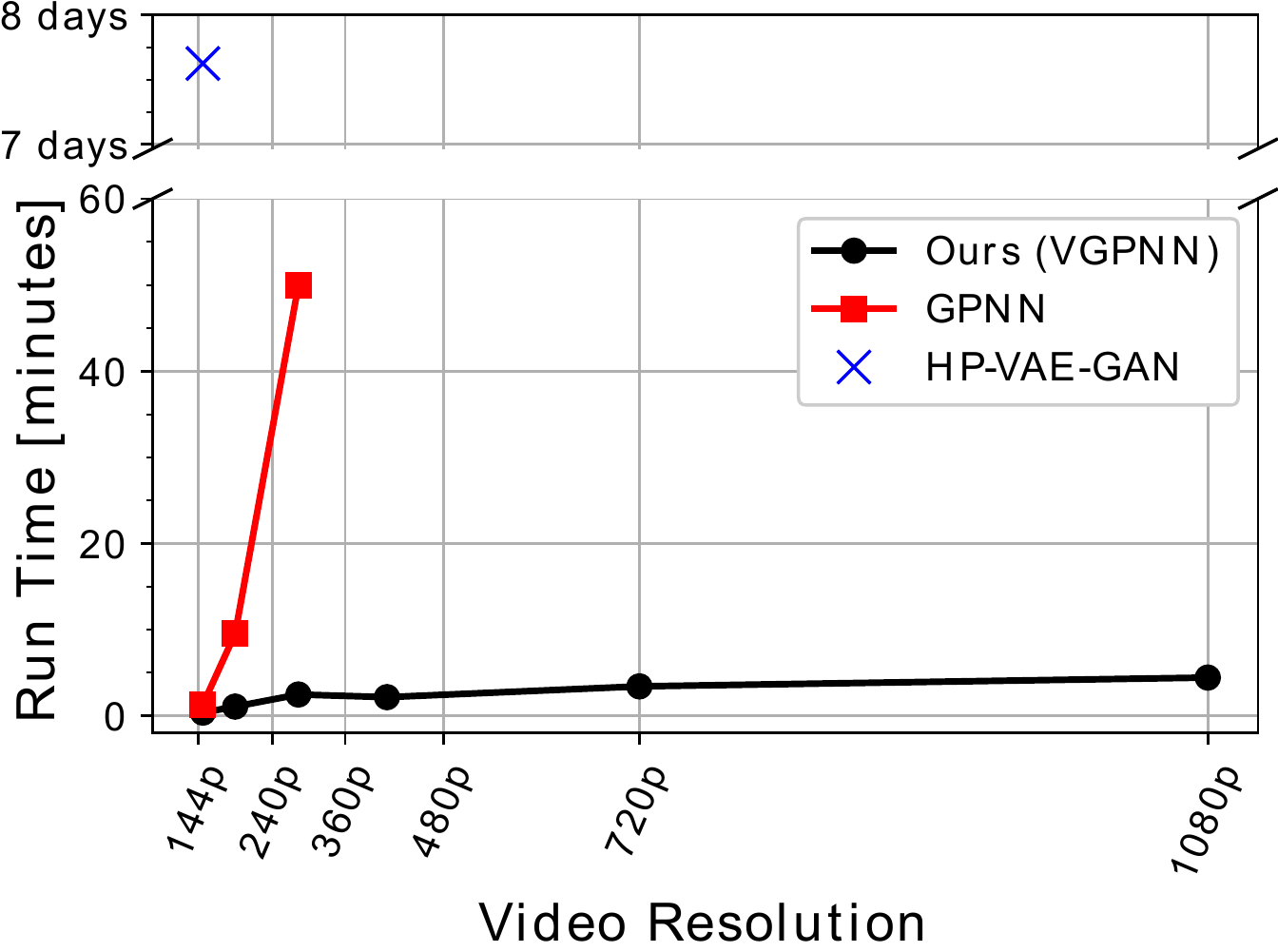}
    \end{tabular}
    \caption{\textbf{Comparing Generation Runtime} between our approach (VGPNN), a na\"ive extension of GPNN~\cite{granot2021drop} from 2D to 3D and \hpvaegan \cite{gur2020hierarchical} on 13-frames videos with different spatial resolutions (X-axis, all have 16:9 aspect ratio).
    \vspace{-15pt}
    }
    \label{fig:runtime_single}
\end{figure}

We use PatchMatch~(\citet{barnes2009patchmatch}) to find the nearest neighbors between $Q$ and $K$ (Fig.~\ref{fig:method_vpnn}b).
The algorithm is implemented on GPU using PyTorch~\cite{NEURIPS2019_9015}, with time complexity $O(n \times d)$ and $O(n)$ additional memory (where $n$ is the video size and $d$ is the patch size). This dramatically reduces both run time and memory footprint used for video generation, making it possible to generate high-resolution videos (see Fig~\ref{fig:runtime_single}).

\vspace{8pt}
\noindent An overview of VPNN module is shown in Fig.~\ref{fig:method_vpnn}.

\vspace{-10pt}
\paragraph{Temporal Diversity and Consistency:}
To enhance the temporal diversity of our samples we set the temporal dimension of the output to be slightly smaller than that of the input video. Thus, motions in different spatial positions in the generated output are taken from different temporal positions in the input video, increasing the overall temporal diversity (see for example the generated dancers in Fig.~\ref{fig:generation_examples} that are not synced).
We also found that the temporal consistency is best preserved in the generated output when the initial noise $z_N$ is randomized for each spatial position, but is the same (replicated) in the temporal dimension.


\vspace{-6pt}
\section{Experimental Results}
\vspace{-4pt}
\label{sec:experiments}

\begin{figure}
  \includegraphics[width=\columnwidth]{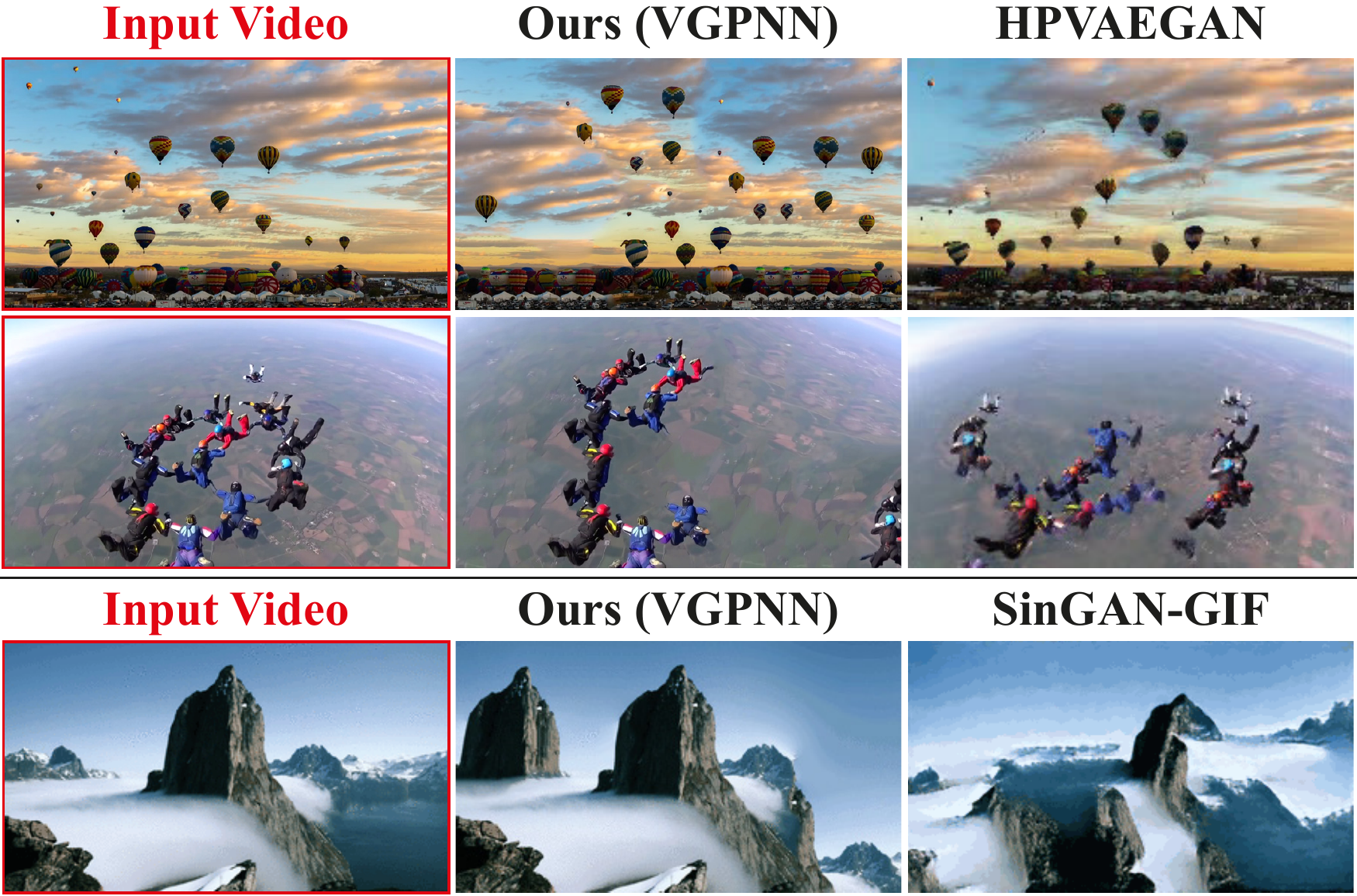}
  \caption{
  \textbf{Comparing Visual Quality} between our generated frames and those of \hpvaegan \cite{gur2020hierarchical} and SinGAN-GIF \cite{arora2021singan} (please \textbf{zoom in} on the frames). Note that our generated frames are sharper and also exhibit more coherent and plausible arrangements of the scene. 
  }
  \vspace{-12pt}
  \label{fig:generation_ours_vs_hpvaegan_singangif}
\end{figure}

We compare our results to those of \hpvaegan~\cite{gur2020hierarchical} and SinGAN-GIF \cite{arora2021singan}, both are methods for diverse video generation from single video. Our results are both qualitatively (Fig.~\ref{fig:generation_ours_vs_hpvaegan_singangif}) and quantitatively (Table~\ref{table:comparison}) superior while reducing the runtime by a factor of $3 \times 10^4$ (from 8 days training on one video to 18 seconds for new generated video). 
While \cite{gur2020hierarchical,arora2021singan} are limited to generated outputs of small resolution ($144{\times}256$), 
the use of efficient PatchMatch algorithm for nearest neighbors search dramatically reduces both run time and memory footprint used for video generation, 
making it possible to generate outputs in the same resolution of the input video (full-HD $1280{\times}1920$)

\begin{table}[t]
\begin{adjustbox}{max width=\columnwidth}
\begin{tabular}{lccc}
\toprule

\textbf{Method} 
&
\textbf{SVFID} 
&
\textbf{Head-on comparison}
&
\textbf{Runtime}   
\\
&
\cite{gur2020hierarchical} $\downarrow$
&
(User study) \textbf{[$\%$]}$\uparrow$
&
$\downarrow$
\\

\midrule

 \hpvaegan \cite{gur2020hierarchical}     & 0.0081  & \multirow{2}{*}{\textbf{67.84}$\pm$1.77} &    7.625 days \\ 
 \textbf{VGPNN} (Ours)  & {\textbf{0.0072}} &    &  \textbf{18} secs \\
 
\midrule
 
 SinGAN-GIF \cite{arora2021singan}  & 0.0119   &  \multirow{2}{*}{\textbf{50.57}$\pm$ 3.27} &  Unpublished \\ 
 \textbf{VGPNN} (Ours) & {\textbf{0.0058}} &  &  \textbf{10} secs \\

\bottomrule

\end{tabular}
\end{adjustbox}
\caption{\textbf{Quantitative Evaluation:} 
SVFID~\cite{gur2020hierarchical} measures the patch statistics similarity between the input video and a generated video. It computes the Fréchet distance between the statistics of the input video and the generated video using pre-computed C3D~\cite{tran2015learning} features (lower is better). 
Note that our generated samples bear more substantial similarity to the input videos.
In a user study, each of 100 AMT participants were asked to judge which of two generated samples was better in terms of sharpness, natural look and coherence. We report the percentage of users who favored our samples over the other.
\vspace{-16pt}
}
\label{table:comparison}

\end{table}

\begin{figure*}
  \includegraphics[width=\textwidth]{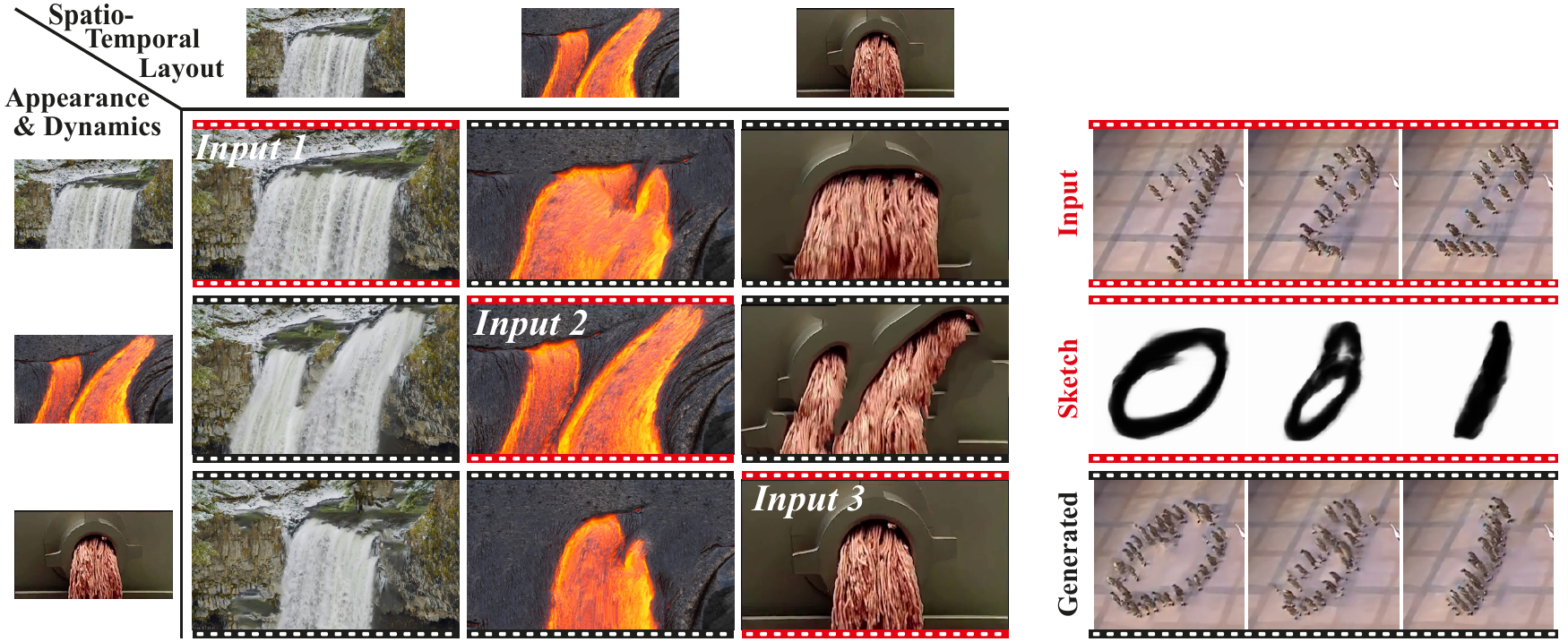}
  \caption{
  \textbf{Video Analogies:} \textit{Left:} an example of video analogies between all pairs of three input videos (red). Each generated video (black) takes the spatio-temporal layout from the input video in its row, and the appearance and dynamics of the input video from its column. \textit{Right:} an example of sketch-to-video -- the generated video (bottom) takes its spatio-temporal layout from the sketch video of morphed MNIST digits (middle) and its appearance and dynamics from the input video of parading soldiers (top). 
  }
  \label{fig:analogies}
\end{figure*}

\vspace{-4pt}
\section{Video analogies}
\label{sec:analogies}
\vspace{-4pt}
Video to video translation methods typically train on large datasets and are either conditioned on human poses or keypoint detection \citep[e.g.][]{wang2018video, wang2019few, mallya2020world, wang2020imaginator}, or require knowledge of a human/animal model \citep[e.g.][]{aberman2020unpaired, chan2019everybody,yang2020transmomo,siarohin2019animating,siarohin2019first,ren2020human,lee2019metapix}. We show that when videos' dynamics are similar in both their motion and semantic context within their video, one can use our framework to transfer the motion and appearance between the two (see Fig.~\ref{fig:analogies}). We term this task ``video analogies'' (inspired by \textit{image analogies} \cite{hertzmann2001image,liao2017visual,benaim2021structural}). More formally, we generate a new video whose spatio-temporal layout is taken from a content video $C$, and overall appearance and dynamics from a style video $S$.

We first extract the ``\textit{dynamic structure}'' of both videos -- the magnitude of the optical flow (extracted via RAFT~\cite{teed2020raft}), quantized into few bins (using k-means). 
We compute the spatio-temporal pyramids of the (i) style video $S$ (ii) dynamic structure of the content video ($\text{dyn}(C)$) (iii) the dynamic structure the style video ($\text{dyn}(S)$). The output video is generated by setting $Q,K,V$ at each level as follows:

$
\begin{array}{@{}l|lll}
     \text{Level} & Q & K & V \\
     \midrule
     \text{N (coarsest)}& \text{dyn}(C)_N & \text{dyn}(S)_N & S_N \\
     \text{n (any other)} & \text{dyn}(C)_n \Vert Q_{n+1}\uparrow & \text{dyn}(S)_n  \Vert S_n & S_n 
\end{array}
$

where $\Vert$ denotes concatenation along the channels axis, and $n$ denote the current level in the pyramid. Note that in the coarsest level, the two videos are only compared by their dynamic structure. In finer levels, the dynamic structure of $C$ (the content video) is used to ``guide'' the output to the desired spatio-temporal layout.
In Fig.~\ref{fig:analogies}left we show snapshots of the analogies of all possible pairs between three videos. In Fig.~\ref{fig:analogies}right we show an example for ``sketch-to-video'' transfer, where the dynamic structure is given by a sketch video instead of an actual video.

\vspace{-6pt}
\section{More Applications}
\vspace{-6pt}
In the \textit{supplementary material} we include examples for spatio-temporal retargeting and conditional video inpainting, as well as further technical details and ablations.

\newpage

{\small
\bibliographystyle{abbrvnat}
\bibliography{vgpnn}
}

\end{document}